\documentclass[10pt, a4paper]{article}

\usepackage{lrec-coling2024} %
\usepackage{microtype}
\usepackage{graphicx}
\usepackage{multirow}
\usepackage{xcolor}  
\usepackage{todonotes}
\usepackage{pifont}
\usepackage{amsmath,amssymb}
\usepackage{bbm}
\usepackage{array}
\usepackage{tabularx}
\usepackage{graphicx}
\usepackage{float} 
\usepackage{subcaption}
\usepackage{epstopdf}
\usepackage{tabularx}
\usepackage[most]{tcolorbox}
\usepackage{booktabs}

\definecolor{darkgreen}{HTML}{228B22}
\definecolor{nblue}{HTML}{377eb8}

\definecolor{ggreen}{HTML}{1b9e77}
\definecolor{oorange}{HTML}{d95f02}
\definecolor{bblue}{HTML}{7570b3}
\definecolor{ppurple}{HTML}{e37fbb}

\definecolor{lgreen}{HTML}{9CD24A}
\definecolor{yyellow}{HTML}{FFD52D}
\definecolor{ggold}{HTML}{E1BC89}
\definecolor{ggray}{HTML}{AAAAAA}

\definecolor{crimson}{HTML}{DC143C}

\newcommand{\xmark}{\textcolor{red}{\ding{55}}}
\newcommand{\cmark}{\textcolor{darkgreen}{\ding{51}}}

\newcommand\Tstrut{\rule{0pt}{2.6ex}}         %
\newcommand\Bstrut{\rule[-0.9ex]{0pt}{0pt}}   %

\newcommand{\model}{XATU}
\newcommand\blfootnote[1]{%
  \begingroup
\renewcommand\thefootnote{}\footnote{#1}%
  \addtocounter{footnote}{-1}%
  \endgroup
  }
\title{XATU: A Fine-grained Instruction-based\\ Benchmark for Explainable Text Updates}

\name{Haopeng Zhang$\dag$, Hayate Iso$\ddagger$, Sairam Gurajada$\ddagger$, Nikita Bhutani$\ddagger$} 

\address{IFM Lab, UC Davis$\dag$, Megagon Labs$\ddagger$ \\
haopeng@ifmlab.org\\
\{hayate, sairam, nikita\}@megagon.ai\\}

\abstract{
   
Text editing is a crucial task of modifying text to better align with user intents. However, existing text editing benchmark datasets contain only coarse-grained instructions and lack explainability, thus resulting in outputs that deviate from the intended changes outlined in the gold reference.
To comprehensively investigate the text editing capabilities of large language models (LLMs), this paper introduces XATU, the first benchmark specifically designed for fine-grained instruction-based explainable text editing. 
XATU considers finer-grained text editing tasks of varying difficulty (simplification, grammar check, fact-check, etc.), incorporating lexical, syntactic, semantic, and knowledge-intensive edit aspects.
To enhance interpretability, we combine LLM-based annotation and human annotation, resulting in a benchmark that includes fine-grained instructions and gold-standard edit explanations. By evaluating existing LLMs against our benchmark, we demonstrate the effectiveness of instruction tuning and the impact of underlying architecture across various editing tasks. Furthermore, extensive experimentation reveals the significant role of explanations in fine-tuning language models for text editing tasks. The benchmark will be open-sourced to support reproduction and facilitate future research at~\url{https://github.com/megagonlabs/xatu}.

 \\ \newline \Keywords{Text Editing, Large Language Model, Language Resources, Explainability}}

\begin{document}

\maketitleabstract

\blfootnote{The work was done when Haopeng Zhang was a research
intern at Megagon Labs}

\section{Introduction}

Text editing is the task of modifying text to better align with user intents. Recent advances in large language models (LLMs) have demonstrated remarkable zero-shot text generation capabilities across a wide range of downstream natural language processing (NLP) tasks like question answering, dialogue, and summarization~\cite{ radford2019language, raffel2020exploring, brown2020language, zhang2022opt,zhang2023extractive}. It has been shown that incorporating instruction tuning~\cite{wei2021finetuned} and reinforcement learning from human feedback (RLHF)~\cite{bai2022training, ouyang2022training} can further enhance a model's ability to align with the user's intent.

Text editing plays a crucial role in real-world text generation applications. While many current text generation models typically employ a one-shot manner, humans often engage in an iterative process when writing text, which entails multiple drafts and revisions~\cite{faltings2020text}. Thus, approaching text generation as an iterative process with successive updates to the text is a more effective way to achieve higher alignment between the model's outputs and the user's intent.

\begin{figure}[!ht]
    \centering
    \includegraphics[width=\linewidth]{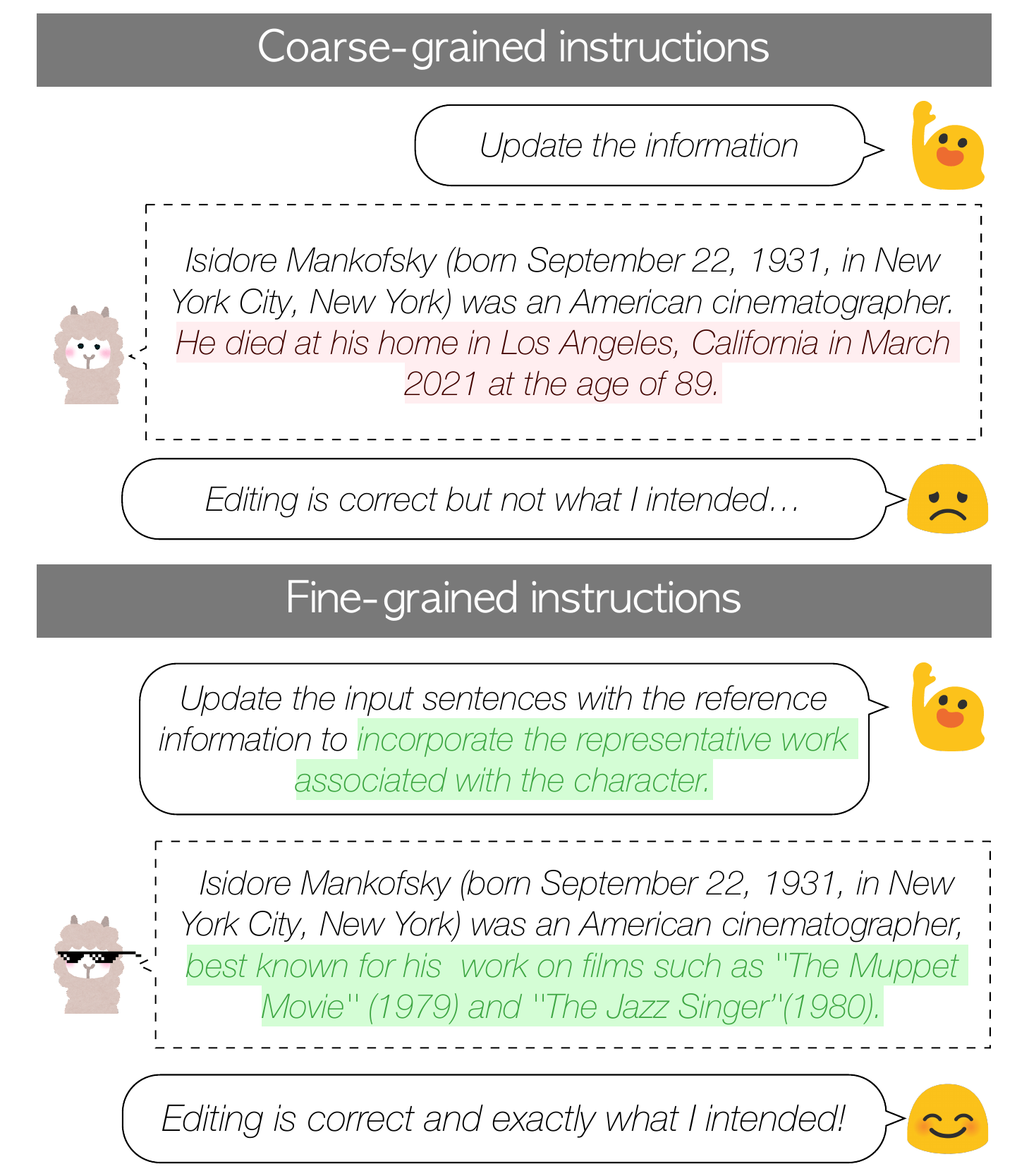}
    \caption{Illustrated examples of coarse- and fine-grained instructions for text editing. LLMs can accurately perform text editing based on coarse-grained instructions, but may not meet the user's intention. In contrast, fine-grained instructions lead to accurate and user-intended text editing.}
    \label{fig:fig1}
\end{figure}
\vspace{-10pt}

\vspace{10pt}
Consequently, \textit{instruction-based text editing} tasks have received growing interest recently. They enable a user to interact with a model with commands to update existing text and achieve desirable text. Typically commands target broad edits such as adding or removing content, or changing the meaning of text~\cite{faltings2020text}. To evaluate models' capabilities for text editing, instruction-based benchmarks such as EditEval~\cite{dwivedi2022editeval} have been proposed recently.

Nevertheless, prior datasets and benchmarks only consider simple commands or instructions, such as `Update the article' or `Expand', resulting in three main limitations. \textbf{(a)} These instructions are often too coarse-grained, to the extent that even humans may struggle to understand the desired intent of the edit, limiting a model's capability to follow instructions. \textbf{(b)} It poses challenges for text editing evaluation since outputs obtained using coarse-grained instructions may not accurately reflect the true editing capabilities of LLM systems. The coarse-grained instructions cannot provide sufficient guidance for the editing system, so the edited output may appear correct but deviate from the intended changes, as shown in Figure~\ref{fig:fig1}. The edited output updates the death information of ``Isidore Mankofsky'' correctly but since the ground truth output updates the information about his famous works. This edit will receive very low evaluation scores since current automatic evaluation methods for text editing systems are mostly based on surface-level overlap with human-written gold references. \textbf{(c)} The current text editing systems lack explainability, which in turn limits the interpretability of the edits made. This lack of interpretability hampers our understanding of the underlying text editing behaviors.

\begin{table}[t]
\small
\centering
\resizebox{\linewidth}{!}{
\begin{tabular}{c|c|c|c|c}
  \hline
  \textbf{Benmark}&
\textbf{Domain}             &  \textbf{Reference} & \textbf{Instruction} & \textbf{Explanation}\Tstrut\Bstrut  \\
  \hline
\shortstack{WikiSemanticIntention \Tstrut\\\cite{yang2017identifying}}& {wiki }                             &           {\xmark }        &  {\xmark} &  {\xmark}\Tstrut\Bstrut\\   %
  \hline
\shortstack{WikiAtomicEdits \Tstrut\\\cite{faruqui2018wikiatomicedits}}& {wiki}                          & {\xmark }                  &           {\xmark }        &  {\xmark} \Tstrut\Bstrut\\  
\hline

  \shortstack{WikiDocEdits \Tstrut\\ \cite{faltings2020text}}& {wiki}                    &           {\cmark }        &  {coarse-grained} &  {\xmark}\Tstrut\Bstrut\\      
  \hline

\shortstack{ITERATER\Tstrut\\\cite{du2022understanding}}& {multiple}                 & {\xmark }                  &           {\xmark }        &  {\xmark} \Tstrut\Bstrut\\   
  \hline

  \shortstack{EditEval\Tstrut\\\cite{dwivedi2022editeval}}& {multiple}                 & {\cmark }                  &           coarse-grained      &  {\xmark} \Tstrut\Bstrut \\   
  \hline
  \shortstack{XATU\Tstrut\\{(ours)}}& {multiple}                   & {\cmark }                  &           fine-grained      &  {\cmark} \Tstrut\Bstrut \\   
  \hline
\end{tabular}}
\caption{A detailed comparison of XATU with existing text editing datasets and benchmarks.}
\label{tab:compare}
\end{table}

To address the aforementioned limitations, we present {\model}, a novel fine-grained instruction-based benchmark designed for e\textbf{\underline{x}}pl\textbf{\underline{a}}inable \textbf{\underline{t}}ext \textbf{\underline{u}}pdates. The benchmark leverages high-quality existing data sources from different tasks to enable automatic evaluation of LLM editing capabilities by incorporating an LLM-in-the-loop annotation process. In comparison to other datasets and benchmarks, {\model} highlights its inclusion of a wider range of diverse topics, fine-grained edit instructions, and corresponding explanation rationales, as shown in Table~\ref{tab:compare}. By utilizing {\model}, we are able to evaluate and compare the performance of several state-of-the-art language models, considering both zero-shot and fine-tuning settings. Experiment results emphasize the significant role of explanations in the fine-tuning process of language models for text editing tasks. Our findings shed light on the importance of incorporating explanation rationales to enhance the performance of language models in text editing, ultimately leading to improved output quality. 
We summarize the contributions of this paper as follows:
\begin{itemize}
    \item We propose XATU, a new benchmark for instruction-based text editing. We collect high-quality data from a set of tasks and datasets for iterative text updates and provide gold fine-grained instructions and explanations with both LLM generation and human annotations. To the best of our knowledge, XATU is the \textbf{first} text editing dataset with \textbf{fine-grained instructions and explanations}.
    \item We conduct a thorough evaluation of existing open and closed large language models on our benchmark with different input formats to test their text editing capabilities. We further investigate the influence of instruction fine-tuning using explanations and fine-grained editing instructions. 
\end{itemize}

\section{Related Work}
\subsection{Text Editing}

Several prior studies have concentrated on the domain of text editing. \citet{guu2018generating} introduced a method that involves generating sentences through the editing of training prototypes. \citet{faruqui2018wikiatomicedits} presented the WikiAtomicEdits dataset, which comprises edits extracted from Wikipedia editing history. Subsequently, \citet{yin2018learning} and \citet{reid2022learning} leveraged Wikipedia editing history to propose the task of learning to represent edits. Furthermore, \citet{iso2020fact} introduced a fact-based text editing task in their work.

Recently, text editing has been approached as an interactive task, leading to the development of command-based editing systems~\cite{faltings2020text} and interactive editing systems~\cite{peerpaper2022}. Researchers have also explored the integration of text editing into interactive self-refinement frameworks for text generation using large language models. \citet{welleck2022generating} introduced a self-corrective learning framework that incorporates a corrector into the language model, enabling self-correction during sequence generation. \citet{akyurek2023rl4f} proposed a reinforcement learning-based approach for generating natural language feedback to correct generation errors. Furthermore, \citet{zhang2023summit} presented an iterative refinement framework for abstractive summarization.

\subsection{Text Editing Datasets}

Previous research has also introduced various datasets and resources that focus on iterative text revisions within specific domains. For instance, datasets presented in ~\citet{yang2017identifying}, ~\citet{faruqui2018wikiatomicedits} and ~\citet{anthonio2020wikihowtoimprove} primarily concentrated on the domain of Wikipedia edit history, while~\citet{spangher2022newsedits} developed a dataset tailored to news articles. Researchers in the field have also proposed more comprehensive benchmarks that cover multiple domains. {IteraTeR}~\cite{du2022understanding} provides iterative tasks from multiple domains, but has only a limited number of tasks, such as fluency, coherence, clarity, style, and meaning change. {EditEval}~\cite{dwivedi2022editeval} is perhaps the closest to our benchmark XATU in that it covers data from multiple domains: Wikipedia, Wikinews, news articles, and arXiv. 

Moreover, it is important to note that only {EditEval} and WikiDocEdits~\cite{faltings2020text} provide short and simple instructions or commands to guide the editing process. In contrast, our benchmark XATU distinguishes itself by including fine-grained instructions for each data instance, which is necessary for assessing the genuine text editing capabilities of different systems. Furthermore, XATU also includes the rationale explanations for each of its instances, which have been proven to be significant for instruction fine-tuning models~\cite{hsieh2023distilling, mukherjee2023orca}.

\section{The {\model} Benchmark}
We introduced the XATU benchmark in this section. We outline the criteria for selecting datasets to construct XATU~(\S\ref{sec:source}), how to annotate fine-grained instructions and explanations for the edits (\S\ref{sec:annotation}), and discuss various use cases (\S\ref{sec:usage}). As shown in Figure~\ref{fig:case}, each instance in XATU comprises five key components: inputs, outputs, optional references, fine-grained instructions, and explanations.

\subsection{Data Source}
\label{sec:source}

In line with the systematic formulation of language style~\cite{dimarco1993computational}, we adopt a similar approach to categorize text editing into four main editing aspects: \textbf{lexical, syntactic, semantic, and knowledge}. Our benchmark XATU aims to serve as a relatively comprehensive benchmark and a holistic evaluation framework for text editing tasks, covering all four aspects of text editing.

\begin{figure}[t]
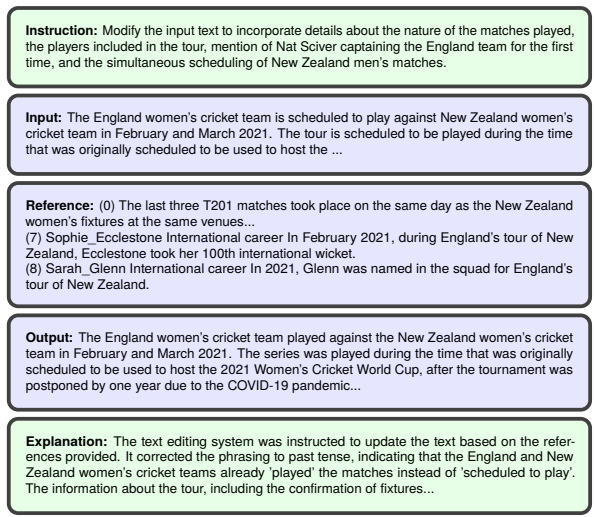

    \centering
    \begin{tcolorbox}[fontupper=\tiny\sffamily, colback=green!10, left=1mm, right=1mm, top=1mm, bottom=1mm]
    \textbf{Instruction:} Modify the input text to incorporate details about the nature of the matches played, the players included in the tour, mention of Nat Sciver captaining the England team for the first time, and the simultaneous scheduling of New Zealand men's matches.
    \end{tcolorbox}\vspace{-1em}
    \begin{tcolorbox}[fontupper=\tiny\sffamily, colback=blue!10, left=1mm, right=1mm, top=1mm, bottom=1mm]
    \textbf{Input:} The England women's cricket team is scheduled to play against New Zealand women's cricket team in February and March 2021. The tour is scheduled to be played during the time that was originally scheduled to be used to host the ...
    \end{tcolorbox}\vspace{-1em}
    \begin{tcolorbox}[fontupper=\tiny\sffamily, colback=blue!10, left=1mm, right=1mm, top=1mm, bottom=1mm]
    \textbf{Reference:} (0) The last three T201 matches took place on the same day as the New Zealand women's fixtures at the same venues...\\
    (7) Sophie\_Ecclestone International career In February 2021, during England's tour of New Zealand, Ecclestone took her 100th international wicket.\\
    (8) Sarah\_Glenn International career In 2021, Glenn was named in the squad for England's tour of New Zealand.
    \end{tcolorbox}\vspace{-1em}
    \begin{tcolorbox}[fontupper=\tiny\sffamily, colback=blue!10, left=1mm, right=1mm, top=1mm, bottom=1mm]
    \textbf{Output:} The England women's cricket team played against the New Zealand women's cricket team in February and March 2021. The series was played during the time that was originally scheduled to be used to host the 2021 Women's Cricket World Cup, after the tournament was postponed by one year due to the COVID-19 pandemic...
    \end{tcolorbox}\vspace{-1em}
    \begin{tcolorbox}[fontupper=\tiny\sffamily, colback=green!10, left=1mm, right=1mm, top=1mm, bottom=1mm]
    \textbf{Explanation:} The text editing system was instructed to update the text based on the references provided. It corrected the phrasing to past tense, indicating that the England and New Zealand women's cricket teams already 'played' the matches instead of 'scheduled to play'. The information about the tour, including the confirmation of fixtures...
    \end{tcolorbox}\vspace{-0.5em}
    \caption{The instance format of the data in XATU benchmark. Data in blue (Input, Output, Reference) are extracted from the corresponding data sources, and data in green (Fine-grained instruction and explanation) are obtained from joint automatic and human annotations.}
    \label{fig:case}
\end{figure}

To ensure comprehensiveness and quality, we meticulously curate high-quality editing data from $4$ distinct downstream NLP tasks, encompassing a total of $1000$ annotated data from $9$ data sources following~\citet{dimarco1993computational}. A summary of the tasks and datasets, along with an overview of the editing aspects covered by each dataset, is presented in Table~\ref{tab:data}. We also transform all datasets within the {\model} benchmark into a consistent format, extracting the input text, gold edits, task type, and optional reference documents from the following data sources, as illustrated in the example in Figure~\ref{fig:case}. This standardized format facilitates evaluation and comparison across different datasets within the benchmark.

\paragraph{Grammar Error Correction} The first task in XATU is to identify and correct errors that don't follow grammar rules. We first incorporate data from JHU FLuency-Extended GUG (JFLEG)~\cite{napoles2017jfleg}, a dataset specifically designed for the task of grammar error correction. JFLEG uses holistic fluency edits to not only correct grammatical
errors but also make the original text more
native sounding. By including this dataset, we aim to assess the basic ability of text editing systems to improve fluency and grammatical accuracy simultaneously.

\paragraph{Simplification} We also include ASSET~\cite{alva2020asset}, a text simplification dataset that provides manually produced simplifications through a wide range of transformation techniques. We aim to evaluate the text editing capabilities of systems when it comes to simplifying complex texts while preserving their essential meaning.

\paragraph{Style Transfer} Text editing also involves transferring the styles of the input sentences, so we include the following three datasets, aiming to evaluate the effectiveness of text editing systems in handling style transfer and bias mitigation tasks.

\begin{table}[t]
\centering
\resizebox{\linewidth}{!}{
\begin{tabular}{l|c|c|c|l}
  \hline
\textbf{Task}             & \textbf{Dataset} & \textbf{Train}&\textbf{Test}&\textbf{Aspect}  \\
  \hline
Grammar& JFLEG   &10  &50       & Lexical, Syntactic            \Tstrut\Bstrut \\   
  \hline
Simplification            & ASSET      &10  &50      &      Lexical, Syntactic   \Tstrut\Bstrut         \\
  \hline

\multirow{3}{*}{Style Transfer} & WNC  &20 &100        & \multirow{3}{*}{Lexical, Semantic}                                    \Tstrut\Bstrut  \\
                          & Wikibias &20  &100            &                           \\
                          & StylePTB &10   &50       &         \Tstrut\Bstrut\\
                            \hline

\multirow{4}{*}{\shortstack{Information\Tstrut\\{Update}}}     & FRUIT      &30  &150                       &    \multirow{4}{*}{Semantic, Knowledge}                               \Tstrut\Bstrut  \\
                          & Evidence&40   &200       &                                   \\
                          & DeFacto &30  &150        &                         \\
                          & Factedit&30   &150       &                   \\
                          \hline

\end{tabular}}
\caption{The detailed datasets included in the XATU benchmark.}
\label{tab:data}
\end{table}

\begin{itemize}
    \item Wiki Neutrality Corpus (WNC)~\cite{pryzant2020automatically} is a collection of original and de-biased sentence pairs mined from Wikipedia edits by carefully filtering based on the editor's comments.
    \item Wikibias~\cite{zhong2021wikibias} is a manually annotated parallel corpus with sentence pairs from Wikipedia editing history. The inclusion of Wikibias complements the mined data from WNC, providing additional annotations to address both sentence-level and token-level biases. 
    \item StylePTB~\cite{lyu2021styleptb} contains paired sentences undergoing various fine-grained stylistic changes and compositions of multiple transfers.
\end{itemize}

\paragraph{Information Update} Our XATU benchmark also includes the following four knowledge-intensive text editing datasets. These datasets are particularly challenging as they require text editing systems to update the input text based on given instructions and external reference evidence. These knowledge-intensive datasets offer a valuable assessment of the systems' capacity to leverage external information for text editing tasks.

\begin{itemize}
    \item FRUIT dataset \citep{logan2021fruit} is a dataset collected by comparing two snapshots of the same Wikipedia article. The reference documents were identified by searching for other Wikipedia articles and human filtering. We include this gold set from the FRUIT dataset in XATU.
    \item Evidence~\cite{thorne2020evidence} is a dataset created using a two-stage distant supervision approach, where evidence is incorporated into masked claims from the FEVER dataset~\cite{thorne2018fever}. Each claim in the dataset is paired with reference {evidence claims} obtained from Wikipedia.
    \item DeFacto~\cite{liu2022improving} dataset consists of document summaries from the XSum dataset~\cite{Narayan2018DontGM}, accompanied by human-corrected versions that rectify factual errors present in the original summaries. For the text editing task, we utilize the {original document} as the reference to evaluate the ability of text editing systems to correct factual errors in the summaries effectively. The inclusion of this dataset allows us to assess the performance of systems in editing text to ensure factual accuracy, an important aspect of text editing in real-world applications.
    \item FactEdit~\cite{iso2020fact} is an editing dataset that incorporates facts sourced from a knowledge base, such as multiple {triples}, as the reference. This dataset is constructed from table-to-text data sources.\\
\end{itemize}

\begin{table}[t]
\small
\centering
\resizebox{0.7\linewidth}{!}{
    \begin{tabular}{clc}
    \toprule
    Difficulty & Dataset & Levenshtein$\downarrow$ \\\midrule
    \multirow{4}{*}{ Easy } & JFLEG & 1.47\\
                            & WNC & 1.58 \\
                            &STYLEPTB & 1.72 \\
                            & Wikibias& 1.92 \\\midrule
    \multirow{3}{*}{ Medium } & Evidence & 3.56 \\
                            & ASSET & 4.72 \\
                            & DeFacto & 4.82 \\\midrule
    \multirow{2}{*}{ Hard } & Factedit & 9.75 \\
                            & FRUIT & 13.62 \\\bottomrule
\end{tabular}    
}
\caption{Average token-level Levenshtein distance between the input and edited output in XATU. }
    \label{tab:diff}
\end{table}

\subsection{Annotation Process}
\label{sec:annotation}

\paragraph{Challenges in Crowdsourcing}
Annotating fine-grained instructions and explanations poses a challenge due to the inherent open-ended nature of the text production task, making it difficult to filter out low-quality workers. Additionally, many workers turn to LLMs for text production tasks due to their prevalence~\cite{veselovsky2023artificial}, but the output quality heavily depends on the user's LLM skills~\cite{zam2023why}. Therefore, relying solely on crowd workers for annotation may not be advisable.

To address these concerns and maintain the quality and reliability of the annotated data, we implemented an LLM-in-the-loop annotation approach, which involves generating candidates of fine-grained instruction and explanations using LLMs and then validating the output with human workers. This allowed us to mitigate the challenges associated with crowd-sourcing and refine our methodology for generating and evaluating fine-grained instructions and editing explanations.

\begin{figure}[t]
    \begin{tcolorbox}[fontupper=\scriptsize\sffamily, fontlower=\scriptsize\sffamily, colback=gray!5, left=1mm, right=1mm, top=1mm, bottom=1mm]
        \textbf{Input: }At least once an episode we see protestors marching around screaming slogans\\
        \textbf{Output: }One or more times an episode protestors are seen walking around yelling blue slogans.
        \tcblower
        \textbf{Input: }{\color{crimson}<del>At least once</del>} an episode {\color{crimson}<del>we see</del>} protestors {\color{crimson}<del>marching</del>} around {\color{crimson}<del>screaming</del>} slogans\\
        \textbf{Output: }{\color{ggreen}<ins>One or more times</ins>} an episode protestors {\color{ggreen}<ins>are seen walking</ins>} around {\color{ggreen}<ins>yelling blue</ins>} slogans {\color{ggreen}<ins>.</ins>}
    \end{tcolorbox}\vspace{-0.5em}
        \caption{Illustrated example of adding HTML tags to the input and output to explicitly indicate the edited portion to LLMs.}
    \label{fig:html_tag}
\end{figure}

\paragraph{Candidate Generation by LLMs}
We generate candidates of fine-grained instructions and explanations for each instance using LLMs~\cite{zhou2023large,honovich-etal-2023-instruction,li2023revisit}.
For the LLMs to generate candidates, we use GPT-4~\cite{openai2023gpt4}.
Through our preliminary prompt engineering, we found that merely asking LLMs to generate the instructions and explanations often resulted in undesired outputs\footnote{More details can be found in the Appendix.}. We found that adding HTML tags to explicitly indicate the edited portion is highly effective, as illustrated in Figure~\ref{fig:html_tag}. Thus, we decided to use this prompting technique to generate instructions and explanations. The full prompt can be found in the Appendix.

Note that our analysis showed no significant differences between coarse and fine-grained instructions for easy tasks (i.e., JFLEG, WNC, STYLEPTB, and WikiBias) since the inputs are relatively short and the edit distance is relatively small. Consequently, we chose to only conduct fine-grained instruction annotation on the more complex information update task ($65\%$ data).

\paragraph{Candidate Validation by Human}
We conducted candidate validation for fine-grained instructions and explanations by employing human evaluation through the Appen platform.\footnote{\url{https://appen.com/}}
Workers were instructed to assess the validity of generated instructions or explanations based on input, output, and optional references. To ensure quality, we selected and manually annotated $20$ random test examples and accepted workers who provided correct answers for over $80\%$ of them.
If a candidate was considered invalid, we repeated the candidate generation process for that instance.

We evaluated the quality of annotations through inter-annotator agreement, which resulted in an overall agreement rate of 80.27\%. We annotated the dataset based on difficulty levels defined in Table~\ref{tab:diff}. A detailed breakdown of the number of instances annotated is provided in Table~\ref{tab:data}. The result of this rigorous process was the creation of the XATU benchmark, containing 1000 high-quality, fine-grained text editing instances.

\subsection{Benchmark Usage}
\label{sec:usage}

Previous work has explored the use of Hamming distance for sentence-level editing task difficulty estimation~\cite{lyu2021styleptb}.  However, as our input and output texts may have different lengths, we employ a token-level Levenshtein distance to measure the relative difficulty of the different datasets in our benchmark. Based on this distance measurement, we categorize the tasks into three levels: easy, medium, and hard. The results of this categorization can be found in Table~\ref{tab:diff}. We notice that information update tasks generally require more editing operations compared to simple lexicon tasks like grammar error correction and neutralization. This highlights the increased complexity and difficulty associated with tasks that involve updating and incorporating external information into the text. To ensure that the benchmark presents a challenging evaluation environment, we deliberately incorporate more examples from demanding tasks.

With the inclusion of fine-grained instructions and corresponding editing rationale explanations, our benchmark offers comprehensive support for a range of traditional text editing tasks like edit representation modeling, automatic editing instruction generation, and editing span prediction. Additionally, the unique structure of XATU enables two new tasks: editing explanation generation and editing evidence retrieval. Our annotated edit explanations can be used to train and assess models aimed at generating coherent and informative explanations for the editing process. Moreover, the information update data that incorporates external reference documents allows for the development and evaluation of models that can effectively retrieve the necessary evidence for text editing tasks.

By providing support for these downstream tasks, our benchmark aims to facilitate advancements in various aspects of text editing, fostering the development of models and techniques that can improve the efficiency and effectiveness of text editing systems.

\section{Experiments}

In this section, we perform a series of text editing experiments using the XATU benchmark, aiming to investigate and address the following research questions:

\begin{itemize}
    \item \textbf{Q1}: What is the actual text editing ability of existing open/closed large language models? 
\end{itemize}   
    By evaluating the performance of LLMs, we aim to assess their effectiveness of instruction-based text editing and understand their limitations and strengths in this context.

\begin{itemize}
    \item \textbf{Q2}: How do fine-grained instructions in XATU differ from the coarse-grained instructions? 
    \end{itemize}
    By comparing the performance of models using fine-grained instructions from XATU, we seek to highlight the impact of instruction granularity on the quality and accuracy of text editing.
    
\begin{itemize}
    
    \item \textbf{Q3}: How do the explanations provided in XATU benefit model tuning under the fine-tuning settings? 
   \end{itemize} 
    
    By examining the performance of models that incorporate the explanations provided in XATU during fine-tuning, we aim to evaluate the effectiveness of these explanations in improving model performance and understanding the role of explanations in the context of text editing.

\begin{table*}[ht]
    \centering
    \resizebox{0.9\textwidth}{!}{

    \begin{tabular}{c|c|c |c| c cc|cccc}
    \hline
    \textbf{Model} & \textbf{Setting} & \textbf{JFLEG} & \textbf{ASSET} & \textbf{WNC} & \textbf{Wikibias}\ & \textbf{PTB} & \textbf{FRUIT} & \textbf{Evidence} & \textbf{DeFacto} & \textbf{Factedit}\Tstrut\Bstrut\\
    \hline
    \multirow{3}{*}{Flan-T5} & coarse & 64.98 & 52.01 & 62.54 & 55.58 \ & 51.05 & 19.06 & 39.13 & 36.77 & 7.98\Tstrut\Bstrut\\
                            & fine & - & - & - & - \ & - & 16.84 & 41.84 & 42.87 & 6.80\\
                            & Exp. & 59.09 & 46.47 & 56.71 & 51.98  & 46.78 & 21.40 & 53.34 & 38.78 & 11.65\Bstrut\\
    \hline
    \multirow{3}{*}{Flan-UL2} & coarse & 64.35 & 50.08 & 59.39 & 51.88 \ & 51.83 & 19.84 & 52.61 & 31.15 & 23.06\Tstrut\Bstrut\\
                            & fine & - & - & - & -\ & - & 43.38 & 68.79 & 52.54 & 35.15\\
                            & Exp. & 83.38 & 66.82 & 86.78 & 79.22  & 74.81 & 38.08 & 74.35 & 52.83 & 29.13\Bstrut\\
    \hline
    \multirow{3}{*}{Alpaca} & coarse & 68.14 & 41.20 & 45.68 & 42.35 \ & 54.14 & 51.75 & 53.65 & 52.58 & 41.08\Tstrut\Bstrut\\
                            & fine & - & - & - & -\ & - & 52.74 & 69.42 & 60.81 & 34.83\\
                            & Exp. & 75.82 & 62.41 & 70.90 & 69.53 & 77.99 & 54.00 & 74.12 & 66.17 & 39.89\Bstrut\\
    \hline
    \multirow{3}{*}{GPT3} & coarse & 50.74 & 30.82 & 32.24 & 34.58\ & 42.63 & 26.47 & 34.12 & 28.42 & 31.46\Tstrut\Bstrut\\
                            & fine & - & - & - & - \ & - & 27.43 & 36.72 & 34.85 & 36.52\\
                            & Exp. & 56.34 & 37.42 & 41.34 & 42.48 & 47.23 & 32.54 & 47.23 & 37.28 & 38.75\Bstrut\\
    \hline
    \multirow{3}{*}{GPT4} & coarse & 70.32 & 56.71 & 64.18 & 58.13 \ & 58.72 & 43.28 & 58.19 & 54.62 & 43.59\Tstrut\Bstrut\\
    & fine & - & - & - & -\ & - & 49.28 & 62.34 & 62.42 & 48.27\\
     & Exp. & \textbf{84.58} & \textbf{73.54} & \textbf{89.23} & \textbf{82.93}  & \textbf{84.39} & \textbf{59.43} & \textbf{81.38} & \textbf{71.38} & \textbf{58.38}\Bstrut\\
    \hline
  
    \end{tabular}
    }
    \caption{Zero-shot text editing results of different LLMs under different prompt settings on XATU benchmark. We omit results on simple datasets that do not require fine-grained instruction. We use coarse, fine, and Exp. to represent text editing with coarse-grained instructions fine-grained instructions, and explanations, respectively.}
    \label{tab:zero}
\end{table*}

\subsection{Experimental Setup}

We include the following baseline LLMs in our experiments:

\begin{itemize}
    \item \textbf{GPT-3} \citep{brown2020language} is a 175B parameter pre-trained decoder-only model. 
    \item \textbf{GPT-4} \citep{openai2023gpt4} is the most recent multimodal large language model created by OpenAI. It was pre-trained to predict the next token and was then fine-tuned with reinforcement learning from human and AI feedback. We evaluate both GPT-3 and GPT-4 through OpenAI's API.\footnote{\url{https://beta.openai.com/}}
    \item \textbf{T5}~\cite{raffel2020exploring} is an encoder-decoder model pre-trained on a multi-task mixture of unsupervised and supervised tasks. 
    \item \textbf{Flan-T5}~\cite{chung2022scaling} is an enhanced version of T5 that has been instruction fine-tuned on a mixture of tasks. 
    \item \textbf{UL2}~\cite{tay2022unifying} is a unified framework for pre-training models with Mixture-of-Denoisers, a pre-training objective that combines diverse pre-training paradigms together. 
    
    \item \textbf{Flan-UL2} is an encoder-decoder model that uses the same configuration as the UL2 model. It was fine-tuned using the "Flan" prompt tuning and dataset collection.
    \item \textbf{LLaMa}~\cite{touvron2023llama} is a collection of foundation language models trained on trillions of tokens. It is an auto-regressive language model built on transformer architecture.

    \item \textbf{Alpaca}~\cite{taori2023alpaca} is an instruction-following language model fine-tuned from the LLaMA 7B model on 52K instruction-following demonstrations. 
    
\end{itemize}

Note that PEER~\cite{peerpaper2022} is another strong text editing baseline according to results in \citet{dwivedi2022editeval}. We didn't include it since the model checkpoint is not publicly available.

For evaluation, we use SARI scores~\cite{Xu-EtAl:2016:TACL}, an $n$-gram based metric commonly used for measuring editing tasks such as simplification~\cite{zhao2018integrating} and sentence fusion~\cite{malmi2019encode}. It has been demonstrated to correlate most closely with human judgment and is computed as:
\begin{align*}
    SARI =(  F1_{add}+  F1_{keep}+ P_{del}) / 3,
\end{align*}
where $F1_{add}, F1_{keep},P_{del}$ represent the F1 scores and precision for add, keep, and delete operations, respectively. We utilize the Hugginface implementation of SARI \footnote{\url{https://huggingface.co/spaces/evaluate-metric/sari}}, where $n=4$. EditEval~\cite{dwivedi2022editeval} includes more n-gram-based valuation metrics, but the results do not demonstrate any significant difference from SARI scores.

\begin{table*}[ht]
    \centering
    \resizebox{0.9\textwidth}{!}{
    \begin{tabular}{c|c|c|c|c cc|cccc}
    \hline
   \textbf{Model} & \textbf{Setting} & \textbf{JFLEG} & \textbf{ASSET} & \textbf{WNC} & \textbf{Wikibias}\ & \textbf{PTB} & \textbf{FRUIT} & \textbf{Evidence} & \textbf{DeFacto} & \textbf{Factedit}\Tstrut\Bstrut\\
    \hline
    \multirow{3}{*}{T5} & coarse & 62.75 & 52.26 & 64.20 & 56.01 \ & 56.59 & 45.27 & 60.03 & 59.07 & 47.65\Tstrut\Bstrut\\
                            & fine & 63.43 & 51.85 & 58.65 & 56.93 \ & 61.07 & 48.42 & 71.56 & 60.91 & 45.80\\
                            & Exp. & 72.73 & 60.23 & 77.39 & 70.76  & 72.22 & 43.83 & 77.24 & 64.23 & 46.14\Bstrut\\
    \hline
    \multirow{3}{*}{Flan-T5} & coarse & 64.07 & 52.71 & 65.04 & 58.86 \ & 63.63 & 50.95 & 62.81 & 59.40 & 45.99\Tstrut\Bstrut\\
                            & fine & 65.39 & 53.00 & 65.11 & 59.56 \ & 63.15 & 53.64 & 76.43 & 66.62 & 47.52\\
                            & Exp. & 79.17 & 69.18 & 84.67 & 75.31   & 75.64 & 52.33 & 85.60 & 71.25 & 47.90\Bstrut\\
    \hline
    \multirow{3}{*}{LLaMA } & coarse & 63.86 & 45.46 & 62.84 & 53.72 \ & 58.87 & 49.18 & 63.69 & 52.09 & 50.25\Tstrut\Bstrut\\
                            & fine & 66.18 & 47.56 & 64.30 & \textbf{61.08} & 60.67 & 53.44 & 82.19 & 69.38 & 52.45\\
                            & Exp. & 83.31 & 70.28 & 91.22 & 84.66 & 84.54 & 54.22 & 86.85 & 79.14 & 55.45\Bstrut\\
    \hline
    \multirow{3}{*}{Alpaca } & coarse & 65.71 & 44.95 & 63.68 & 55.77 \ & 63.62 & 49.18 & 64.13 & 56.73 & 46.73\Tstrut\Bstrut\\
                            & fine & 69.64 & 47.14 & 62.78 & 50.57\ & 61.18 & 51.81 & 83.48 & \textbf{73.69} & 46.35\\
                            & Exp. & 83.52 & 70.83 & 87.93 & 75.85 & 83.56 & 58.33 & 88.41 & 77.91 & 47.45\Bstrut\\

    \hline
    \multirow{3}{*}{UL2 } & coarse & 66.61 & \textbf{54.55} & 69.82 & \textbf{60.45} \ & 61.86 & 51.49 & 70.65 & \textbf{59.21} & 58.18\Tstrut\Bstrut\\
                            & fine & 71.22 & \textbf{54.84} & 71.19 & 56.04 \  & 67.27 & 56.58 & 84.37 & 70.27 & 56.06\\
                            & Exp. & 87.81 & 78.22 & 91.79 & 82.24 & \textbf{88.10} & 54.64 & 90.54 & 78.65 & 56.44\Bstrut\\
    \hline
        \multirow{3}{*}{Flan-UL2 } & coarse & \textbf{68.03} & 52.34 & \textbf{73.93} & 57.24 \ & \textbf{73.12} & \textbf{51.81} & \textbf{71.80} & 58.01 & \textbf{58.72}\Tstrut\Bstrut\\
                            & fine & \textbf{71.84} & 52.59 & \textbf{75.16} & 58.09 \  & \textbf{69.38} & \textbf{63.91} & \textbf{86.17} & 73.56 & \textbf{60.64}\\
                            & Exp. & \textbf{90.44} & \textbf{79.84} & \textbf{94.68} & \textbf{86.23} & 86.06 & \textbf{59.46} & \textbf{91.71} & \textbf{82.84} & \textbf{60.76}\Bstrut\\
    \hline
    \end{tabular}
    }
    
    \caption{Instruction fine-tuning results of different LLMs under different prompt settings on XATU. We use coarse, fine, and Exp. to represent text editing with coarse-grained instructions fine-grained instructions, and explanations, respectively.}
    \label{tab:fine-tune}
\end{table*}

\paragraph{Implementation Details}
We use the Huggingface implementation of all models with LoRA~\cite{hu2022lora} for the fine-tuning experiments. We instruction-tuned all models for $15$ epochs with $200$ training examples at a learning rate of $1e-5$ and tested on the original $1000$ test set, as shown in Table~\ref{tab:data}. We keep a small training data size as \cite{zhou2023lima} found that alignment doesn't require too much data. We use LoRA~\cite{hu2022lora} to instruction fine-tune all models for $15$ epochs with $200$ additional training examples at a learning rate of $1e-5$.

\subsection{Zero-shot Results}
The zero-shot evaluation results on the XATU benchmark are presented in Table~\ref{tab:zero}, which highlights several important observations. 

\textbf{GPT-4 demonstrates exceptional zero-shot editing performance}, surpassing all other large language models by a significant margin across all datasets. Alpaca emerges as the second-strongest model, but there is still a considerable performance gap when compared to GPT-4. On the other hand, models without instruction tuning, such as GPT-3, struggle to follow the instructions effectively, resulting in relatively low SARI scores. This emphasizes the importance of instruction tuning in enhancing the text editing performance of language models.

In addition, we find that \textbf{almost all models exhibit improvements when guided by the fine-grained instructions} provided in XATU, as opposed to simple and coarse-grained instructions. This underscores the significance of fine-grained instructions in enhancing the performance and accuracy of text editing tasks. Notably, the inclusion of explanations as guidance during the text editing process leads to further improvements across all models. 

Furthermore, our findings reveal that \textbf{the underlying architecture (encoder-decoder vs. decoder-only) of language models significantly impacts the performance of different types of text editing tasks}. We can see that Flan-T5 and Flan-UL2 exhibit biased performance, excelling in more straightforward tasks like style transfer but facing challenges in more complex tasks like information updates. In contrast, the Alpaca model performs well on information update tasks but achieves lower scores in simple neutralization tasks. We attribute this difference in performance to the \textit{underlying architecture} of the foundation models: Flan-T5 and Flan-UL2 are both encoder-decoder models, while Alpaca is built upon the LLaMa decoder-only model. The results demonstrate that the encoder's ability to understand and represent the input text is more important for lexical and syntactic editing, while the decoder's capacity to generate new and relevant text appears to be more influential for knowledge-intensive tasks.

\subsection{Fine-tuning Results}

After fine-tuning all models using different prompting formats, we evaluate the fine-tuned models on XATU, and the results are presented in Table~\ref{tab:fine-tune}. A notable observation is the significant performance improvement compared to the zero-shot results across all datasets and under all settings. This demonstrates \textbf{the effectiveness of instruction fine-tuning, even with a limited number of examples} for text editing tasks ($200$ in this case).

Among all the instruction fine-tuned models, Flan-UL2 consistently exhibits the strongest performance across all three settings. This indicates its strong adaptation capability during the fine-tuning process.
Furthermore, we observe that the fine-grained instructions in XATU lead to larger performance improvements in tasks that have fine-grained instructions (e.g. FRUIT, DeFacto), while the improvements are comparatively smaller in simpler tasks that lack fine-grained instructions (e.g. WNC, PTB) in the fine-tuning data. 

Comparing T5 and UL2 with their instruction-tuned versions (Flan-T5 and Flan-UL2), we observe that \textbf{the instruction-tuned versions of LLM consistently outperform their base models}. This is especially evident when the models are guided with fine-grained instructions or explanations. These results further emphasize the effectiveness of the instruction alignment ability that these models acquire through instruction fine-tuning.

Overall, the fine-tuning process significantly improves the performance of the models on text editing tasks. The presence of fine-grained instructions and explanations further enhances the performance.

\subsection{Discussion}
\begin{figure}[ht]
    \centering
    \includegraphics[width=0.48\textwidth]{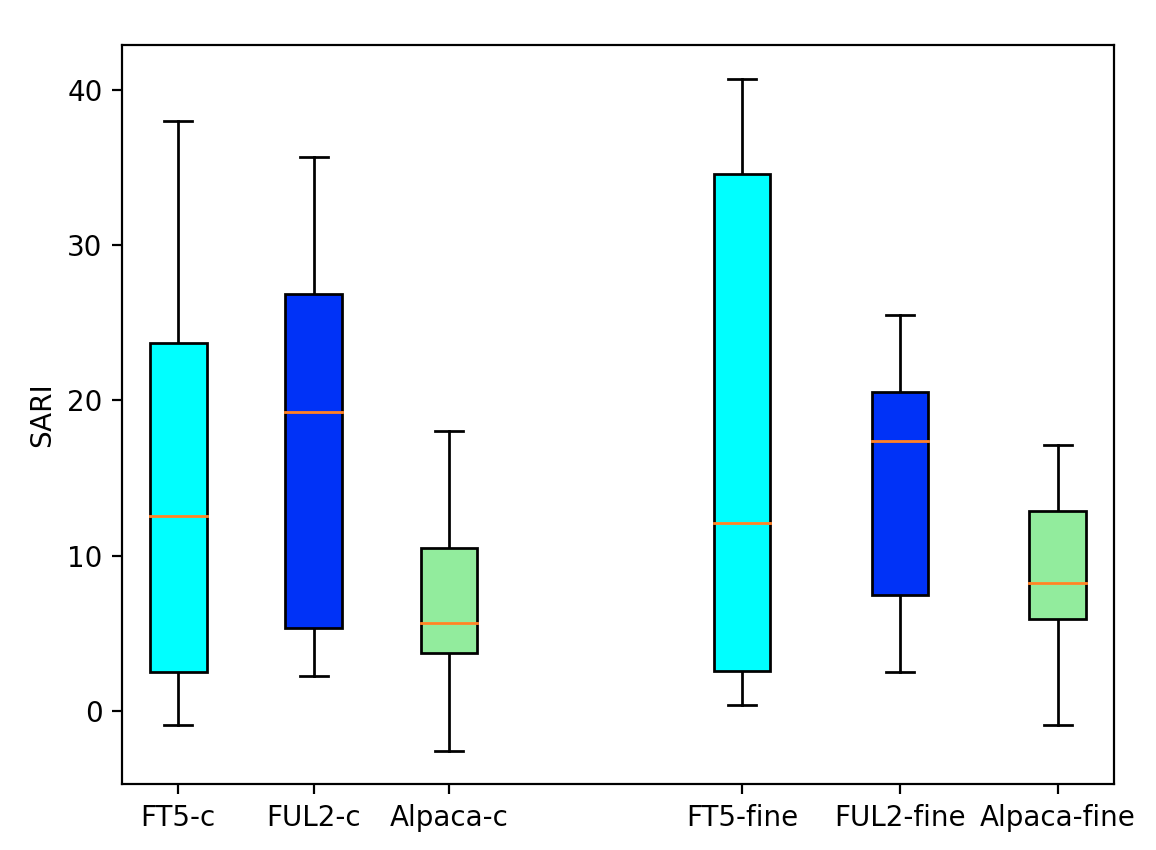}
    \caption{Fine-tuning with fine-grained instructions (-fine) vs. coarse instructions (-c).}
    \label{fig:fine}
\end{figure}
\begin{figure}[ht]
    \centering
    \includegraphics[width=0.48\textwidth]{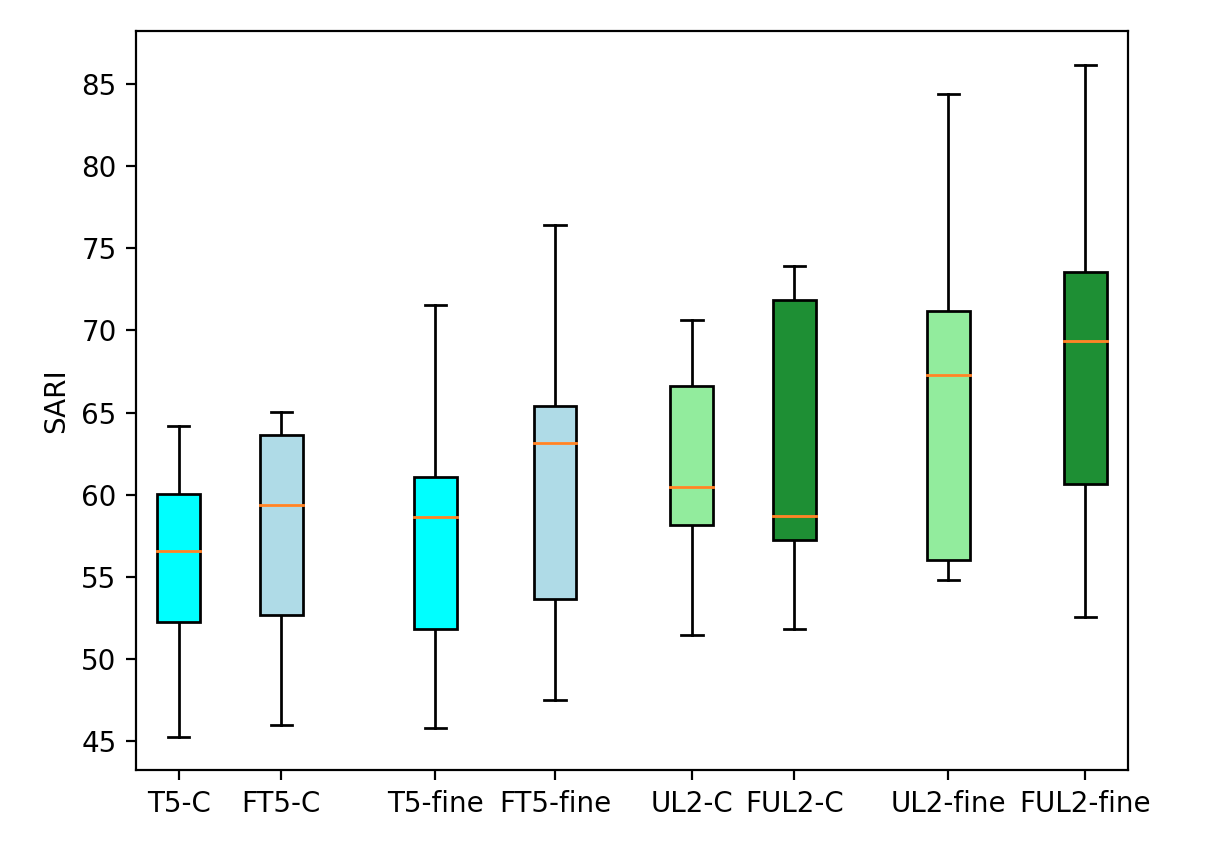}
    \caption{Boxplot comparing instruction-tuned LLMs (Flan-xx) vs. pre-trained counterparts with fine-grained (-fine) and coarse instructions (-c).}
    \label{fig:flan}
\end{figure}

As illustrated in Figure~\ref{fig:fine}, we compare the results of three models after instruction fine-tuning with fine-grained instructions in XATU vs. coarse-grained instructions. We observe that fine-grained instruction models consistently outperform their coarse-grained counterparts, indicating the effectiveness of detailed instruction in improving instruction following and text editing capabilities. Across both coarse-grained and fine-grained instruction settings, we find that the Flan-UL2 model achieves the largest average improvements. On the other hand, Alpaca demonstrates superior robustness across all text editing tasks compared to Flan-T5 and Flan-UL2. 

As depicted in Figure~\ref{fig:flan}, the Flan-UL2 model consistently outperforms the UL2 model across all settings. Similarly, the Flan-T5 model performs better than the base T5 model. These results highlight the effectiveness of instruction tuning in improving the performance of language models across different settings. The instruction-tuning procedure allows the models to better understand and follow the provided instructions, resulting in more accurate and appropriate text edits. We also notice edits generated with coarse-grained instructions show better robustness across all tasks.

\section{Conclusion}
This paper introduces XATU, the first benchmark for explainable text updates with fine-grained instructions. XATU is a diverse benchmark covering a wide range of topics and text types and leverages high-quality data sources from various existing sources. 
We compare existing open and closed instruction-tuned language models under both the zero-shot and fine-tuning settings and reveal their capabilities to edit text and follow instructions.
By releasing the benchmark to the community, we hope to stimulate further research in developing instruction-based text editing models and even potentially interactive text generation systems. 

\section{Acknowledgements}

We thank the anonymous reviewers for their valuable reviews. Additionally, we appreciate Tom Mitchell and the entire team at Megagon Labs for insightful discussions.

\section{Bibliographical References}\label{sec:reference}

\bibliographystyle{lrec-coling2024-natbib}
\bibliography{lrec-coling2024-example}

\begin{thebibliography}{51}
\expandafter\ifx\csname natexlab\endcsname\relax\def\natexlab#1{#1}\fi

\bibitem[{Aky{\"u}rek et~al.(2023)Aky{\"u}rek, Aky{\"u}rek, Madaan, Kalyan, Clark, Wijaya, and Tandon}]{akyurek2023rl4f}
Afra~Feyza Aky{\"u}rek, Ekin Aky{\"u}rek, Aman Madaan, Ashwin Kalyan, Peter Clark, Derry Wijaya, and Niket Tandon. 2023.
\newblock Rl4f: Generating natural language feedback with reinforcement learning for repairing model outputs.
\newblock \emph{arXiv preprint arXiv:2305.08844}.

\bibitem[{Alva-Manchego et~al.(2020)Alva-Manchego, Martin, Bordes, Scarton, Sagot, and Specia}]{alva2020asset}
Fernando Alva-Manchego, Louis Martin, Antoine Bordes, Carolina Scarton, Beno{\^\i}t Sagot, and Lucia Specia. 2020.
\newblock \href {https://doi.org/10.18653/v1/2020.acl-main.424} {{ASSET}: {A} dataset for tuning and evaluation of sentence simplification models with multiple rewriting transformations}.
\newblock In \emph{Proceedings of the 58th Annual Meeting of the Association for Computational Linguistics}, pages 4668--4679, Online. Association for Computational Linguistics.

\bibitem[{Anthonio et~al.(2020)Anthonio, Bhat, and Roth}]{anthonio2020wikihowtoimprove}
Talita Anthonio, Irshad Bhat, and Michael Roth. 2020.
\newblock \href {https://aclanthology.org/2020.lrec-1.702} {wiki{H}ow{T}o{I}mprove: A resource and analyses on edits in instructional texts}.
\newblock In \emph{Proceedings of the Twelfth Language Resources and Evaluation Conference}, pages 5721--5729, Marseille, France. European Language Resources Association.

\bibitem[{Bai et~al.(2022)Bai, Jones, Ndousse, Askell, Chen, DasSarma, Drain, Fort, Ganguli, Henighan et~al.}]{bai2022training}
Yuntao Bai, Andy Jones, Kamal Ndousse, Amanda Askell, Anna Chen, Nova DasSarma, Dawn Drain, Stanislav Fort, Deep Ganguli, Tom Henighan, et~al. 2022.
\newblock Training a helpful and harmless assistant with reinforcement learning from human feedback.
\newblock \emph{arXiv preprint arXiv:2204.05862}.

\bibitem[{Brown et~al.(2020)Brown, Mann, Ryder, Subbiah, Kaplan, Dhariwal, Neelakantan, Shyam, Sastry, Askell, Agarwal, Herbert-Voss, Krueger, Henighan, Child, Ramesh, Ziegler, Wu, Winter, Hesse, Chen, Sigler, Litwin, Gray, Chess, Clark, Berner, McCandlish, Radford, Sutskever, and Amodei}]{brown2020language}
Tom Brown, Benjamin Mann, Nick Ryder, Melanie Subbiah, Jared~D Kaplan, Prafulla Dhariwal, Arvind Neelakantan, Pranav Shyam, Girish Sastry, Amanda Askell, Sandhini Agarwal, Ariel Herbert-Voss, Gretchen Krueger, Tom Henighan, Rewon Child, Aditya Ramesh, Daniel Ziegler, Jeffrey Wu, Clemens Winter, Chris Hesse, Mark Chen, Eric Sigler, Mateusz Litwin, Scott Gray, Benjamin Chess, Jack Clark, Christopher Berner, Sam McCandlish, Alec Radford, Ilya Sutskever, and Dario Amodei. 2020.
\newblock \href {https://proceedings.neurips.cc/paper/2020/file/1457c0d6bfcb4967418bfb8ac142f64a-Paper.pdf} {Language models are few-shot learners}.
\newblock In \emph{Advances in Neural Information Processing Systems}, volume~33, pages 1877--1901. Curran Associates, Inc.

\bibitem[{Chung et~al.(2022)Chung, Hou, Longpre, Zoph, Tay, Fedus, Li, Wang, Dehghani, Brahma et~al.}]{chung2022scaling}
Hyung~Won Chung, Le~Hou, Shayne Longpre, Barret Zoph, Yi~Tay, William Fedus, Eric Li, Xuezhi Wang, Mostafa Dehghani, Siddhartha Brahma, et~al. 2022.
\newblock Scaling instruction-finetuned language models.
\newblock \emph{arXiv preprint arXiv:2210.11416}.

\bibitem[{DiMarco and Hirst(1993)}]{dimarco1993computational}
Chrysanne DiMarco and Graeme Hirst. 1993.
\newblock A computational theory of goal-directed style in syntax.
\newblock \emph{Computational Linguistics}, 19(3):451--500.

\bibitem[{Du et~al.(2022)Du, Raheja, Kumar, Kim, Lopez, and Kang}]{du2022understanding}
Wanyu Du, Vipul Raheja, Dhruv Kumar, Zae~Myung Kim, Melissa Lopez, and Dongyeop Kang. 2022.
\newblock \href {https://doi.org/10.18653/v1/2022.acl-long.250} {Understanding iterative revision from human-written text}.
\newblock In \emph{Proceedings of the 60th Annual Meeting of the Association for Computational Linguistics (Volume 1: Long Papers)}, pages 3573--3590, Dublin, Ireland. Association for Computational Linguistics.

\bibitem[{Dwivedi-Yu et~al.(2022)Dwivedi-Yu, Schick, Jiang, Lomeli, Lewis, Izacard, Grave, Riedel, and Petroni}]{dwivedi2022editeval}
Jane Dwivedi-Yu, Timo Schick, Zhengbao Jiang, Maria Lomeli, Patrick Lewis, Gautier Izacard, Edouard Grave, Sebastian Riedel, and Fabio Petroni. 2022.
\newblock Editeval: An instruction-based benchmark for text improvements.
\newblock \emph{arXiv preprint arXiv:2209.13331}.

\bibitem[{Faltings et~al.(2021)Faltings, Galley, Hintz, Brockett, Quirk, Gao, and Dolan}]{faltings2020text}
Felix Faltings, Michel Galley, Gerold Hintz, Chris Brockett, Chris Quirk, Jianfeng Gao, and Bill Dolan. 2021.
\newblock \href {https://doi.org/10.18653/v1/2021.naacl-main.414} {Text editing by command}.
\newblock In \emph{Proceedings of the 2021 Conference of the North American Chapter of the Association for Computational Linguistics: Human Language Technologies}, pages 5259--5274, Online. Association for Computational Linguistics.

\bibitem[{Faruqui et~al.(2018)Faruqui, Pavlick, Tenney, and Das}]{faruqui2018wikiatomicedits}
Manaal Faruqui, Ellie Pavlick, Ian Tenney, and Dipanjan Das. 2018.
\newblock \href {https://doi.org/10.18653/v1/D18-1028} {{W}iki{A}tomic{E}dits: A multilingual corpus of {W}ikipedia edits for modeling language and discourse}.
\newblock In \emph{Proceedings of the 2018 Conference on Empirical Methods in Natural Language Processing}, pages 305--315, Brussels, Belgium. Association for Computational Linguistics.

\bibitem[{Guu et~al.(2018)Guu, Hashimoto, Oren, and Liang}]{guu2018generating}
Kelvin Guu, Tatsunori~B. Hashimoto, Yonatan Oren, and Percy Liang. 2018.
\newblock \href {https://doi.org/10.1162/tacl_a_00030} {Generating sentences by editing prototypes}.
\newblock \emph{Transactions of the Association for Computational Linguistics}, 6:437--450.

\bibitem[{Honovich et~al.(2023)Honovich, Shaham, Bowman, and Levy}]{honovich-etal-2023-instruction}
Or~Honovich, Uri Shaham, Samuel~R. Bowman, and Omer Levy. 2023.
\newblock \href {https://doi.org/10.18653/v1/2023.acl-long.108} {Instruction induction: From few examples to natural language task descriptions}.
\newblock In \emph{Proceedings of the 61st Annual Meeting of the Association for Computational Linguistics (Volume 1: Long Papers)}, pages 1935--1952, Toronto, Canada. Association for Computational Linguistics.

\bibitem[{Hsieh et~al.(2023)Hsieh, Li, Yeh, Nakhost, Fujii, Ratner, Krishna, Lee, and Pfister}]{hsieh2023distilling}
Cheng-Yu Hsieh, Chun-Liang Li, Chih-Kuan Yeh, Hootan Nakhost, Yasuhisa Fujii, Alexander Ratner, Ranjay Krishna, Chen-Yu Lee, and Tomas Pfister. 2023.
\newblock Distilling step-by-step! outperforming larger language models with less training data and smaller model sizes.
\newblock \emph{arXiv preprint arXiv:2305.02301}.

\bibitem[{Hu et~al.(2022)Hu, yelong shen, Wallis, Allen-Zhu, Li, Wang, Wang, and Chen}]{hu2022lora}
Edward~J Hu, yelong shen, Phillip Wallis, Zeyuan Allen-Zhu, Yuanzhi Li, Shean Wang, Lu~Wang, and Weizhu Chen. 2022.
\newblock \href {https://openreview.net/forum?id=nZeVKeeFYf9} {Lo{RA}: Low-rank adaptation of large language models}.
\newblock In \emph{International Conference on Learning Representations}.

\bibitem[{Iso et~al.(2020)Iso, Qiao, and Li}]{iso2020fact}
Hayate Iso, Chao Qiao, and Hang Li. 2020.
\newblock \href {https://doi.org/10.18653/v1/2020.acl-main.17} {{F}act-based {T}ext {E}diting}.
\newblock In \emph{Proceedings of the 58th Annual Meeting of the Association for Computational Linguistics}, pages 171--182, Online. Association for Computational Linguistics.

\bibitem[{Iv et~al.(2022)Iv, Passos, Singh, and Chang}]{logan2021fruit}
Robert Iv, Alexandre Passos, Sameer Singh, and Ming-Wei Chang. 2022.
\newblock \href {https://doi.org/10.18653/v1/2022.naacl-main.269} {{FRUIT}: Faithfully reflecting updated information in text}.
\newblock In \emph{Proceedings of the 2022 Conference of the North American Chapter of the Association for Computational Linguistics: Human Language Technologies}, pages 3670--3686, Seattle, United States. Association for Computational Linguistics.

\bibitem[{Li et~al.(2023)Li, Zhang, and Zhang}]{li2023revisit}
Zizhong Li, Haopeng Zhang, and Jiawei Zhang. 2023.
\newblock A revisit of fake news dataset with augmented fact-checking by chatgpt.
\newblock \emph{arXiv preprint arXiv:2312.11870}.

\bibitem[{Liu et~al.(2022)Liu, Deb, Teruel, Halfaker, Radev, and Awadallah}]{liu2022improving}
Yixin Liu, Budhaditya Deb, Milagro Teruel, Aaron Halfaker, Dragomir Radev, and Ahmed~H Awadallah. 2022.
\newblock On improving summarization factual consistency from natural language feedback.
\newblock \emph{arXiv preprint arXiv:2212.09968}.

\bibitem[{Lyu et~al.(2021)Lyu, Liang, Pham, Hovy, P{\'o}czos, Salakhutdinov, and Morency}]{lyu2021styleptb}
Yiwei Lyu, Paul~Pu Liang, Hai Pham, Eduard Hovy, Barnab{\'a}s P{\'o}czos, Ruslan Salakhutdinov, and Louis-Philippe Morency. 2021.
\newblock \href {https://doi.org/10.18653/v1/2021.naacl-main.171} {{S}tyle{PTB}: A compositional benchmark for fine-grained controllable text style transfer}.
\newblock In \emph{Proceedings of the 2021 Conference of the North American Chapter of the Association for Computational Linguistics: Human Language Technologies}, pages 2116--2138, Online. Association for Computational Linguistics.

\bibitem[{Malmi et~al.(2019)Malmi, Krause, Rothe, Mirylenka, and Severyn}]{malmi2019encode}
Eric Malmi, Sebastian Krause, Sascha Rothe, Daniil Mirylenka, and Aliaksei Severyn. 2019.
\newblock \href {https://doi.org/10.18653/v1/D19-1510} {Encode, tag, realize: High-precision text editing}.
\newblock In \emph{Proceedings of the 2019 Conference on Empirical Methods in Natural Language Processing and the 9th International Joint Conference on Natural Language Processing (EMNLP-IJCNLP)}, pages 5054--5065, Hong Kong, China. Association for Computational Linguistics.

\bibitem[{Mukherjee et~al.(2023)Mukherjee, Mitra, Jawahar, Agarwal, Palangi, and Awadallah}]{mukherjee2023orca}
Subhabrata Mukherjee, Arindam Mitra, Ganesh Jawahar, Sahaj Agarwal, Hamid Palangi, and Ahmed Awadallah. 2023.
\newblock Orca: Progressive learning from complex explanation traces of gpt-4.
\newblock \emph{arXiv preprint arXiv:2306.02707}.

\bibitem[{Napoles et~al.(2017)Napoles, Sakaguchi, and Tetreault}]{napoles2017jfleg}
Courtney Napoles, Keisuke Sakaguchi, and Joel Tetreault. 2017.
\newblock \href {https://aclanthology.org/E17-2037} {{JFLEG}: A fluency corpus and benchmark for grammatical error correction}.
\newblock In \emph{Proceedings of the 15th Conference of the {E}uropean Chapter of the Association for Computational Linguistics: Volume 2, Short Papers}, pages 229--234, Valencia, Spain. Association for Computational Linguistics.

\bibitem[{Narayan et~al.(2018)Narayan, Cohen, and Lapata}]{Narayan2018DontGM}
Shashi Narayan, Shay~B. Cohen, and Mirella Lapata. 2018.
\newblock \href {https://doi.org/10.18653/v1/D18-1206} {Don{'}t give me the details, just the summary! topic-aware convolutional neural networks for extreme summarization}.
\newblock In \emph{Proceedings of the 2018 Conference on Empirical Methods in Natural Language Processing}, pages 1797--1807, Brussels, Belgium. Association for Computational Linguistics.

\bibitem[{OpenAI(2023)}]{openai2023gpt4}
OpenAI. 2023.
\newblock \href {http://arxiv.org/abs/2303.08774} {Gpt-4 technical report}.

\bibitem[{Ouyang et~al.(2022)Ouyang, Wu, Jiang, Almeida, Wainwright, Mishkin, Zhang, Agarwal, Slama, Ray et~al.}]{ouyang2022training}
Long Ouyang, Jeff Wu, Xu~Jiang, Diogo Almeida, Carroll~L Wainwright, Pamela Mishkin, Chong Zhang, Sandhini Agarwal, Katarina Slama, Alex Ray, et~al. 2022.
\newblock Training language models to follow instructions with human feedback.
\newblock \emph{arXiv preprint arXiv:2203.02155}.

\bibitem[{Pryzant et~al.(2020)Pryzant, Martinez, Dass, Kurohashi, Jurafsky, and Yang}]{pryzant2020automatically}
Reid Pryzant, Richard~Diehl Martinez, Nathan Dass, Sadao Kurohashi, Dan Jurafsky, and Diyi Yang. 2020.
\newblock Automatically neutralizing subjective bias in text.
\newblock In \emph{Proceedings of the aaai conference on artificial intelligence}, volume~34, pages 480--489.

\bibitem[{Radford et~al.(2019)Radford, Wu, Child, Luan, Amodei, and Sutskever}]{radford2019language}
Alec Radford, Jeff Wu, Rewon Child, David Luan, Dario Amodei, and Ilya Sutskever. 2019.
\newblock \href {https://cdn.openai.com/better-language-models/language_models_are_unsupervised_multitask_learners.pdf} {Language models are unsupervised multitask learners}.
\newblock Technical report, Open AI.

\bibitem[{Raffel et~al.(2020)Raffel, Shazeer, Roberts, Lee, Narang, Matena, Zhou, Li, Liu et~al.}]{raffel2020exploring}
Colin Raffel, Noam Shazeer, Adam Roberts, Katherine Lee, Sharan Narang, Michael Matena, Yanqi Zhou, Wei Li, Peter~J Liu, et~al. 2020.
\newblock Exploring the limits of transfer learning with a unified text-to-text transformer.
\newblock \emph{J. Mach. Learn. Res.}, 21(140):1--67.

\bibitem[{Reid and Neubig(2022)}]{reid2022learning}
Machel Reid and Graham Neubig. 2022.
\newblock \href {https://aclanthology.org/2022.findings-emnlp.280} {Learning to model editing processes}.
\newblock In \emph{Findings of the Association for Computational Linguistics: EMNLP 2022}, pages 3822--3832, Abu Dhabi, United Arab Emirates. Association for Computational Linguistics.

\bibitem[{Schick et~al.(2023)Schick, Yu, Jiang, Petroni, Lewis, Izacard, You, Nalmpantis, Grave, and Riedel}]{peerpaper2022}
Timo Schick, Jane~A. Yu, Zhengbao Jiang, Fabio Petroni, Patrick Lewis, Gautier Izacard, Qingfei You, Christoforos Nalmpantis, Edouard Grave, and Sebastian Riedel. 2023.
\newblock \href {https://openreview.net/forum?id=KbYevcLjnc} {{PEER}: A collaborative language model}.
\newblock In \emph{The Eleventh International Conference on Learning Representations}.

\bibitem[{Spangher et~al.(2022)Spangher, Ren, May, and Peng}]{spangher2022newsedits}
Alexander Spangher, Xiang Ren, Jonathan May, and Nanyun Peng. 2022.
\newblock \href {https://doi.org/10.18653/v1/2022.naacl-main.10} {{N}ews{E}dits: A news article revision dataset and a novel document-level reasoning challenge}.
\newblock In \emph{Proceedings of the 2022 Conference of the North American Chapter of the Association for Computational Linguistics: Human Language Technologies}, pages 127--157, Seattle, United States. Association for Computational Linguistics.

\bibitem[{Taori et~al.(2023)Taori, Gulrajani, Zhang, Dubois, Li, Guestrin, Liang, and Hashimoto}]{taori2023alpaca}
Rohan Taori, Ishaan Gulrajani, Tianyi Zhang, Yann Dubois, Xuechen Li, Carlos Guestrin, Percy Liang, and Tatsunori~B Hashimoto. 2023.
\newblock Alpaca: A strong, replicable instruction-following model.
\newblock \emph{Stanford Center for Research on Foundation Models. https://crfm. stanford. edu/2023/03/13/alpaca. html}, 3(6):7.

\bibitem[{Tay et~al.(2023)Tay, Dehghani, Tran, Garcia, Wei, Wang, Chung, Bahri, Schuster, Zheng, Zhou, Houlsby, and Metzler}]{tay2022unifying}
Yi~Tay, Mostafa Dehghani, Vinh~Q. Tran, Xavier Garcia, Jason Wei, Xuezhi Wang, Hyung~Won Chung, Dara Bahri, Tal Schuster, Steven Zheng, Denny Zhou, Neil Houlsby, and Donald Metzler. 2023.
\newblock \href {https://openreview.net/forum?id=6ruVLB727MC} {{UL}2: Unifying language learning paradigms}.
\newblock In \emph{The Eleventh International Conference on Learning Representations}.

\bibitem[{Thorne and Vlachos(2021)}]{thorne2020evidence}
James Thorne and Andreas Vlachos. 2021.
\newblock \href {https://doi.org/10.18653/v1/2021.acl-long.256} {Evidence-based factual error correction}.
\newblock In \emph{Proceedings of the 59th Annual Meeting of the Association for Computational Linguistics and the 11th International Joint Conference on Natural Language Processing (Volume 1: Long Papers)}, pages 3298--3309, Online. Association for Computational Linguistics.

\bibitem[{Thorne et~al.(2018)Thorne, Vlachos, Christodoulopoulos, and Mittal}]{thorne2018fever}
James Thorne, Andreas Vlachos, Christos Christodoulopoulos, and Arpit Mittal. 2018.
\newblock \href {https://doi.org/10.18653/v1/N18-1074} {{FEVER}: a large-scale dataset for fact extraction and {VER}ification}.
\newblock In \emph{Proceedings of the 2018 Conference of the North {A}merican Chapter of the Association for Computational Linguistics: Human Language Technologies, Volume 1 (Long Papers)}, pages 809--819, New Orleans, Louisiana. Association for Computational Linguistics.

\bibitem[{Touvron et~al.(2023)Touvron, Lavril, Izacard, Martinet, Lachaux, Lacroix, Rozi{\`e}re, Goyal, Hambro, Azhar et~al.}]{touvron2023llama}
Hugo Touvron, Thibaut Lavril, Gautier Izacard, Xavier Martinet, Marie-Anne Lachaux, Timoth{\'e}e Lacroix, Baptiste Rozi{\`e}re, Naman Goyal, Eric Hambro, Faisal Azhar, et~al. 2023.
\newblock Llama: Open and efficient foundation language models.
\newblock \emph{arXiv preprint arXiv:2302.13971}.

\bibitem[{Veselovsky et~al.(2023)Veselovsky, Ribeiro, and West}]{veselovsky2023artificial}
Veniamin Veselovsky, Manoel~Horta Ribeiro, and Robert West. 2023.
\newblock \href {http://arxiv.org/abs/2306.07899} {Artificial artificial artificial intelligence: Crowd workers widely use large language models for text production tasks}.

\bibitem[{Wei et~al.(2022)Wei, Bosma, Zhao, Guu, Yu, Lester, Du, Dai, and Le}]{wei2021finetuned}
Jason Wei, Maarten Bosma, Vincent Zhao, Kelvin Guu, Adams~Wei Yu, Brian Lester, Nan Du, Andrew~M. Dai, and Quoc~V Le. 2022.
\newblock \href {https://openreview.net/forum?id=gEZrGCozdqR} {Finetuned language models are zero-shot learners}.
\newblock In \emph{International Conference on Learning Representations}.

\bibitem[{Welleck et~al.(2023)Welleck, Lu, West, Brahman, Shen, Khashabi, and Choi}]{welleck2022generating}
Sean Welleck, Ximing Lu, Peter West, Faeze Brahman, Tianxiao Shen, Daniel Khashabi, and Yejin Choi. 2023.
\newblock \href {https://openreview.net/forum?id=hH36JeQZDaO} {Generating sequences by learning to self-correct}.
\newblock In \emph{The Eleventh International Conference on Learning Representations}.

\bibitem[{Xu et~al.(2016)Xu, Napoles, Pavlick, Chen, and Callison-Burch}]{Xu-EtAl:2016:TACL}
Wei Xu, Courtney Napoles, Ellie Pavlick, Quanze Chen, and Chris Callison-Burch. 2016.
\newblock \href {https://cocoxu.github.io/publications/tacl2016-smt-simplification.pdf} {Optimizing statistical machine translation for text simplification}.
\newblock \emph{Transactions of the Association for Computational Linguistics}, 4:401--415.

\bibitem[{Yang et~al.(2017)Yang, Halfaker, Kraut, and Hovy}]{yang2017identifying}
Diyi Yang, Aaron Halfaker, Robert Kraut, and Eduard Hovy. 2017.
\newblock Identifying semantic edit intentions from revisions in wikipedia.
\newblock In \emph{Proceedings of the 2017 Conference on Empirical Methods in Natural Language Processing}, pages 2000--2010.

\bibitem[{Yin et~al.(2019)Yin, Neubig, Allamanis, Brockschmidt, and Gaunt}]{yin2018learning}
Pengcheng Yin, Graham Neubig, Miltiadis Allamanis, Marc Brockschmidt, and Alexander~L. Gaunt. 2019.
\newblock \href {https://openreview.net/forum?id=BJl6AjC5F7} {Learning to represent edits}.
\newblock In \emph{International Conference on Learning Representations}.

\bibitem[{Zamfirescu-Pereira et~al.(2023)Zamfirescu-Pereira, Wong, Hartmann, and Yang}]{zam2023why}
J.D. Zamfirescu-Pereira, Richmond~Y. Wong, Bjoern Hartmann, and Qian Yang. 2023.
\newblock \href {https://doi.org/10.1145/3544548.3581388} {Why johnny can’t prompt: How non-ai experts try (and fail) to design llm prompts}.
\newblock In \emph{Proceedings of the 2023 CHI Conference on Human Factors in Computing Systems}, CHI '23, New York, NY, USA. Association for Computing Machinery.

\bibitem[{Zhang et~al.(2023{\natexlab{a}})Zhang, Liu, and Zhang}]{zhang2023extractive}
Haopeng Zhang, Xiao Liu, and Jiawei Zhang. 2023{\natexlab{a}}.
\newblock \href {https://doi.org/10.18653/v1/2023.findings-emnlp.214} {Extractive summarization via {C}hat{GPT} for faithful summary generation}.
\newblock In \emph{Findings of the Association for Computational Linguistics: EMNLP 2023}, pages 3270--3278, Singapore. Association for Computational Linguistics.

\bibitem[{Zhang et~al.(2023{\natexlab{b}})Zhang, Liu, and Zhang}]{zhang2023summit}
Haopeng Zhang, Xiao Liu, and Jiawei Zhang. 2023{\natexlab{b}}.
\newblock \href {https://doi.org/10.18653/v1/2023.findings-emnlp.714} {{S}umm{I}t: Iterative text summarization via {C}hat{GPT}}.
\newblock In \emph{Findings of the Association for Computational Linguistics: EMNLP 2023}, pages 10644--10657, Singapore. Association for Computational Linguistics.

\bibitem[{Zhang et~al.(2022)Zhang, Roller, Goyal, Artetxe, Chen, Chen, Dewan, Diab, Li, Lin, Mihaylov, Ott, Shleifer, Shuster, Simig, Koura, Sridhar, Wang, and Zettlemoyer}]{zhang2022opt}
Susan Zhang, Stephen Roller, Naman Goyal, Mikel Artetxe, Moya Chen, Shuohui Chen, Christopher Dewan, Mona Diab, Xian Li, Xi~Victoria Lin, Todor Mihaylov, Myle Ott, Sam Shleifer, Kurt Shuster, Daniel Simig, Punit~Singh Koura, Anjali Sridhar, Tianlu Wang, and Luke Zettlemoyer. 2022.
\newblock \href {https://doi.org/10.48550/ARXIV.2205.01068} {Opt: Open pre-trained transformer language models}.

\bibitem[{Zhao et~al.(2018)Zhao, Meng, He, Saptono, and Parmanto}]{zhao2018integrating}
Sanqiang Zhao, Rui Meng, Daqing He, Andi Saptono, and Bambang Parmanto. 2018.
\newblock \href {https://doi.org/10.18653/v1/D18-1355} {Integrating transformer and paraphrase rules for sentence simplification}.
\newblock In \emph{Proceedings of the 2018 Conference on Empirical Methods in Natural Language Processing}, pages 3164--3173, Brussels, Belgium. Association for Computational Linguistics.

\bibitem[{Zhong(2021)}]{zhong2021wikibias}
Yang Zhong. 2021.
\newblock \emph{WIKIBIAS: Detecting Multi-Span Subjective Biases in Language}.
\newblock Ph.D. thesis, The Ohio State University.

\bibitem[{Zhou et~al.(2023{\natexlab{a}})Zhou, Liu, Xu, Iyer, Sun, Mao, Ma, Efrat, Yu, Yu et~al.}]{zhou2023lima}
Chunting Zhou, Pengfei Liu, Puxin Xu, Srini Iyer, Jiao Sun, Yuning Mao, Xuezhe Ma, Avia Efrat, Ping Yu, Lili Yu, et~al. 2023{\natexlab{a}}.
\newblock Lima: Less is more for alignment.
\newblock \emph{arXiv preprint arXiv:2305.11206}.

\bibitem[{Zhou et~al.(2023{\natexlab{b}})Zhou, Muresanu, Han, Paster, Pitis, Chan, and Ba}]{zhou2023large}
Yongchao Zhou, Andrei~Ioan Muresanu, Ziwen Han, Keiran Paster, Silviu Pitis, Harris Chan, and Jimmy Ba. 2023{\natexlab{b}}.
\newblock \href {https://openreview.net/forum?id=92gvk82DE-} {Large language models are human-level prompt engineers}.
\newblock In \emph{The Eleventh International Conference on Learning Representations}.

\end{thebibliography}

\clearpage
\appendix

\section{Data Annotation}
\label{appen:annotation}
\begin{figure*}[!tb]
    \centering
    \includegraphics[width=\textwidth]{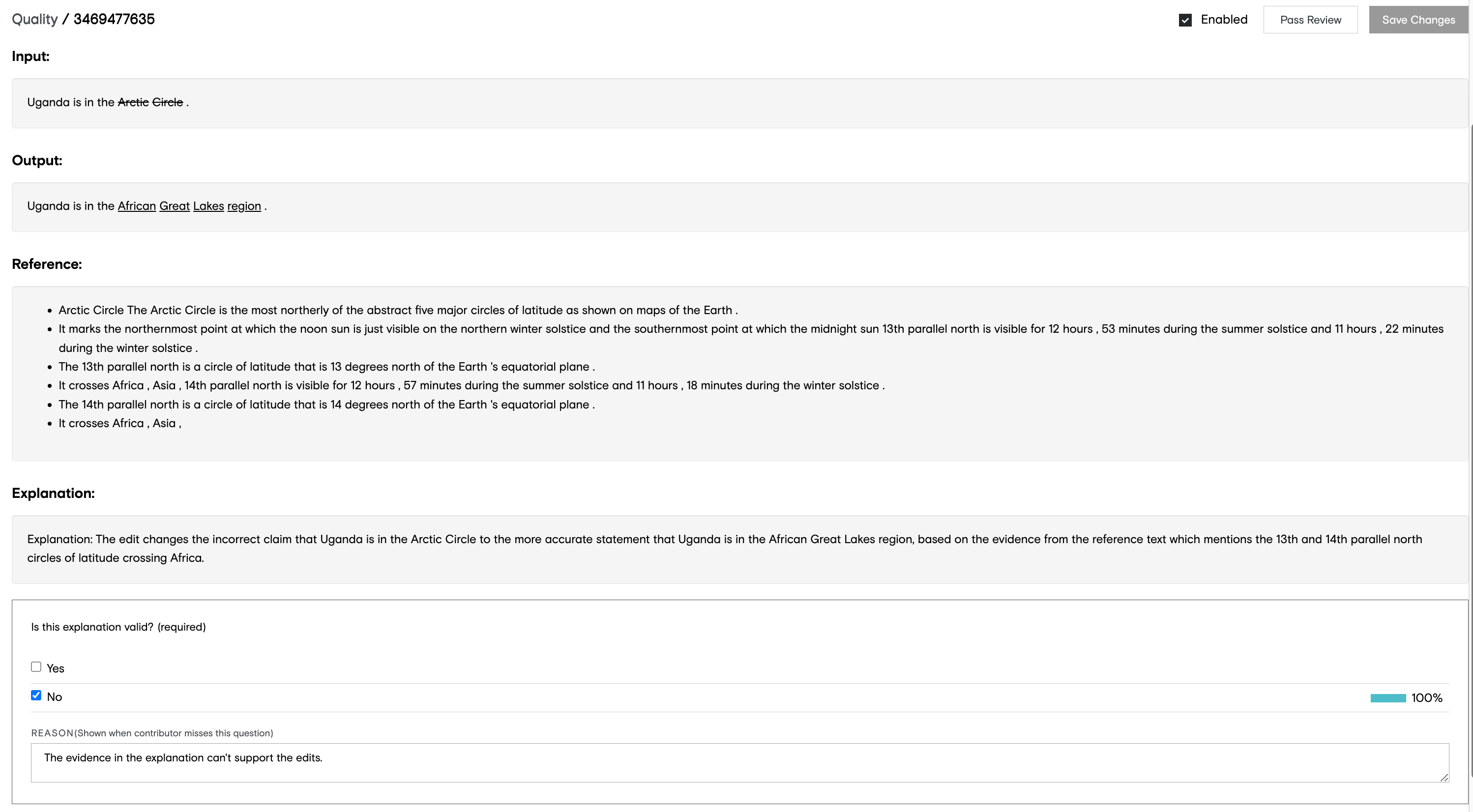}
    \caption{The interface used for annotation.}
    \label{fig:interface}
\end{figure*}

The human annotations were conducted on the Appen\footnote{\url{https://appen.com/}} platform, an example user interface is shown in Figure~\ref{fig:interface}.

The prompts used to generate the explanation are:

\begin{tcolorbox}[fontupper=\ttfamily, title={\small Prompts to generate fine-grained instruction}]
\scriptsize
[no prose]\\

\# Task:\\
Your task is to write a detailed instruction that enables the AI assistant to edit the original text into a revised text based on the references. The instruction must not cause information leakage about the revised text.\\

\# Original text:\\
\{\{input\}\}\\

\# Reference:\\
\{\{reference\}\}\\

\# Revised text:\\
\{\{output\}\}\\

\# Instruction:
\end{tcolorbox}

\begin{tcolorbox}[fontupper=\ttfamily, title={\small Prompts to generate explanation}]
\scriptsize
[no prose]\\

\# Task:\\
Your task is to provide a two-sentence explanation of the edits made based on the instruction and reference by comparing the original and revised texts.\\

\# Instruction:\\
\{\{instruction\}\}\\

\# Original text:\\
\{\{input\}\}\\

\# Reference:\\
\{\{reference\}\}\\

\# Revised text:\\
\{\{output\}\}\\

\# Explanation:
\end{tcolorbox}

\section{Implementation Details}
\label{appen:imple}
The detailed instruction fine-tuning parameters are summarized in Table~\ref{tab:parameter}. We use LoRA~\cite{hu2022lora} to instruction fine-tune all models for $15$ epochs with $200$ additional training examples at a learning rate of $1e-5$.

\begin{table}[htbp]
\small
    \centering
\begin{tabular}{c|c}
\toprule
 $\mathtt{lora}_{\mathtt{r}}$
 & 8 \\
$\mathtt{lora}_{\alpha}$
 & 32 \\
 $\mathtt{num\_epochs}$& 15 \\
 lr & 1e-5\\
 $\mathtt{lora}_{\mathtt{dropout}}$ & 0.1 \\
 $\mathtt{seed}$ & 634 \\
 $\mathtt{max\_length}$ & 1024 \\
\bottomrule
\end{tabular}    
\caption{Hyper-parameters used in the fine-tuning experiments. }
    \label{tab:parameter}
\end{table}

The prompts used to generate the text editing and fine-tuning are:

\begin{tcolorbox}[fontupper=\ttfamily, title={\small Prompts for Text Editing}]
\scriptsize
[no prose]\\

Below is an instruction that describes a task, along with an input text paired with a reference and an explanation that provides further context. Please edit the input text based on the instructions, the reference, and the explanation. Your response should only include the edited output.\\

\# Instruction:\\
\{\{instruction\}\}\\

\# Input:\\
\{\{input\}\}\\

\# Reference:\\
\{\{reference\}\}\\

\# Explanation:\\
\{\{explanation\}\}\\

\# Response:
\end{tcolorbox}

\end{document}